\documentclass[runningheads]{llncs}

\makeatletter
\AtBeginDocument{
  \pagestyle{headings}

  \def\myAuthorText{Do Minh Duc et al.} 
  
  \def\myTitleText{Auto-Prompting with Retrieval Guidance for Frame
Detection in Logistics}

  \renewcommand{\@evenhead}{%
    \small\normalfont
    \myAuthorText      
    \hfill             
    \hspace{2em}       
    \thepage           
  }

  \renewcommand{\@oddhead}{%
    \small\normalfont
    \myTitleText       
    \hfill             
    \hspace{2em}       
    \thepage           
  }
}
\makeatother

\usepackage[T1]{fontenc}
\usepackage{graphicx}

\usepackage[utf8]{inputenc}   
\usepackage[vietnamese, english]{babel}

\usepackage{times}
\usepackage{latexsym}
\usepackage{framed}

\usepackage{microtype}

\usepackage{inconsolata}

\usepackage{graphicx}
\usepackage{adjustbox}
\usepackage{tabularx}
\usepackage{makecell}
\addto\captionsvietnamese{
  \renewcommand{\tablename}{Table}
  \renewcommand{\figurename}{Figure}
  \renewcommand{\abstractname}{Abstract}
}
\usepackage{hyperref}

\begin{document}
\title{Auto-Prompting with Retrieval Guidance for Frame Detection in Logistics}
%
%

\author{
  \textbf{Do Minh Duc\textsuperscript{1}}\orcidID{0009-0008-4116-4067},
  \textbf{Quan Xuan Truong\textsuperscript{1}}\orcidID{0009-0001-4911-3394},
  \textbf{Nguyen Tat Dat \textsuperscript{1}}\orcidID{0009-0005-7422-2663},
  \textbf{Nguyen Van Vinh\textsuperscript{1, *}\orcidID{0009-0004-0912-6328}}
  \institute{\textsuperscript{1}Faculty of Information Technology, VNU University of Engineering and Technology, Hanoi, Vietnam}
\\[4pt]
  \small{
    \href{mailto:dominhduc@vnu.edu.vn}{dominhduc@vnu.edu.vn},
    \href{mailto:xuantruongvnuet@gmail.com}{xuantruongvnuet@gmail.com},
    \href{mailto:nguyentatdat2811@gmail.com}{nguyentatdat2811@gmail.com},
    \href{mailto:vinhnv@vnu.edu.vn}{vinhnv@vnu.edu.vn}
  }
}
\authorrunning{Do Minh Duc et al.}


\maketitle   
\footnotetext[1]{\textsuperscript{*} Corresponding author}

\begin{abstract}
Prompt engineering plays a critical role in adapting large language models (LLMs) to complex reasoning and labeling tasks without the need for extensive fine-tuning. In this paper, we propose a novel prompt optimization pipeline for frame detection in logistics texts, combining retrieval-augmented generation (RAG), few-shot prompting, chain-of-thought (CoT) reasoning, and automatic CoT synthesis (Auto-CoT) to generate highly effective task-specific prompts. Central to our approach is an LLM-based prompt optimizer agent that iteratively refines the prompts using retrieved examples, performance feedback, and internal self-evaluation. Our framework is evaluated on a real-world logistics text annotation task, where reasoning accuracy and labeling efficiency are critical. Experimental results show that the optimized prompts—particularly those enhanced via Auto-CoT and RAG—improve real-world inference accuracy by up to 15\% compared to baseline zero-shot or static prompts. The system demonstrates consistent improvements across multiple LLMs, including GPT-4o, Qwen 2.5 (72B), and LLaMA 3.1 (70B), validating its generalizability and practical value. These findings suggest that structured prompt optimization is a viable alternative to full fine-tuning, offering scalable solutions for deploying LLMs in domain-specific NLP applications such as logistics.
\end{abstract}

\section{Introduction}

The recent breakthroughs in large language models (LLMs), such as ChatGPT~\cite{Achiam2023GPT4TR}, Gemini~\cite{Reid2024Gemini1U}, and other foundation models, have significantly reshaped how businesses approach automation. These models demonstrate remarkable abilities in understanding natural language, reasoning over complex instructions, and adapting across diverse use cases without extensive task-specific fine-tuning~\cite{Bender2021OnTD}. As a result, AI-driven automation is being increasingly deployed across industries to streamline workflows, reduce operational costs, and augment human decision-making~\cite{Wang2025LLMsFS}.

In the logistics sector, where large volumes of informal text such as shipment updates, incident reports, or customer inquiries must be processed daily, LLMs have the potential to automate previously manual workflows~\cite{Jamee2025EnhancingSC}. A key task in this context is \textbf{frame detection}, which involves identifying structured semantic events and their roles (e.g., “delayed delivery,” “handover responsibility”) from unstructured short messages. This can be modeled as a multi-level classification task, where each message may carry nested frame labels representing process stage, location, and agent responsibility. However, real-world deployments face two major challenges: (1) extremely low-resource conditions, with only 1–2 labeled examples per frame type~\cite{Jiang2023LowResourceTC}, and (2) domain-specific terminology that LLMs are not exposed to during pretraining~\cite{Gu2020DomainSpecificLM}.

Building on the prompting paradigm introduced by McCann et al.~\cite{McCann2018TheNL}, LLMs are now widely used via prompt-based interfaces. Prompting has been enhanced through few-shot learning~\cite{Brown2020LanguageMA}, instruction tuning~\cite{Ouyang2022TrainingLM}, and zero-shot reasoning~\cite{Kojima2022LargeLM}. However, prompting remains sensitive: small variations in wording, ordering, or formatting can result in large differences in output quality~\cite{Wang2023ASO}, making it difficult for non-expert users to construct reliable prompts.

To overcome these limitations, \textbf{Automatic Prompt Optimization (APO)} methods have been proposed to refine prompts in a black-box setting, without requiring model fine-tuning. Although effective in general domains, APO remains underexplored in domain-specific, low-resource contexts such as logistics.

In this work, we introduce \textbf{Auto-Prompting with Retrieval Guidance}, a retrieval-enhanced APO approach tailored to frame detection in logistics. Our method:
\begin{itemize}
    \item Maintains a library of prompt components, including structured examples and instruction templates;
    \item Uses semantic similarity retrieval to dynamically select relevant components per input;
    \item Composes optimized prompts without human intervention;
    \item Achieves more than 80\% accuracy on a proprietary logistics dataset, outperforming zero-shot and manually engineered prompts.
\end{itemize}

In summary, by bridging APO with retrieval-based contextualization~\cite{Lewis2020RetrievalAugmentedGF}, our method reduces prompt engineering burden, improves robustness, and facilitates practical deployment of LLMs for structured frame detection in specialized domains. Conceptually, the framework can be viewed as an iterative search in the prompt space, where retrieval narrows contextual uncertainty and self-evaluation acts as a feedback signal approximating gradient-free optimization. This provides a theoretical justification for its effectiveness in black-box settings.

\section{Related Work}
\subsection{LLMs in Domain-Specific and Low-Resource Scenarios}
The unprecedented capabilities of LLMs like GPT-4 \cite{Achiam2023GPT4TR} have spurred their adoption in various specialized, high-stakes domains that often suffer from data scarcity. In medicine, models are being developed to assist with clinical note summarization and diagnostic reasoning \cite{Singhal2022LargeLM}. Similarly, in the legal sector, LLMs are being explored for contract analysis and legal question-answering, where domain-specific terminology and nuanced reasoning are critical. The financial domain has also seen the emergence of models like FinGPT, trained on financial data to handle tasks such as sentiment analysis and report generation \cite{Yang2023FinGPTOF}.

The logistics sector presents similar challenges: a reliance on proprietary data formats, a specialized vocabulary, and the need to process vast streams of unstructured text. Recent surveys suggest that LLMs hold significant potential for automating logistics workflows, such as shipment tracking and incident reporting\cite{Wang2023ASO}. Our work addresses a core task in this domain: frame detection and many labels we only have 1 or 2 samples.

\subsection{Automatic Prompt Optimization}
The effectiveness of LLMs is heavily dependent on the quality of the input prompt, a phenomenon that has given rise to the field of "prompt engineering." However, manual prompt crafting is often brittle and requires significant expertise \cite{Jiang2019HowCW,Zhao2021CalibrateBU}. To address this, Automatic Prompt Optimization (APO) has emerged as a key research direction.

APO techniques can be broadly categorized. Gradient-based methods, such as prompt tuning \cite{Lester2021ThePO}, learn continuous "soft prompts" that are prepended to the input embedding. However, these methods typically require a moderate amount of labeled data for effective training. In our case, the dataset is extremely low-resource, with only 1–2 labeled examples per label, making conventional fine-tuning or even prompt tuning impractical. Given the large number of distinct labels and the limited supervision available, we instead leverage the generative capacity of LLMs themselves to optimize prompts in a data-efficient manner. A seminal work in this area, Automatic Prompt Engineer (APE) \cite{Zhou2022LargeLM}, uses an LLM to generate and search for instruction candidates. More recent approaches have framed this as a reinforcement learning problem. For instance, Pryzant et al \cite{Pryzant2023AutomaticPO} employ "gradient-free" exploration, using an editor model to propose prompt edits and a reward model to evaluate their efficacy. These methods are particularly appealing for low-resource settings, as they minimize dependence on labeled examples while still enabling prompt adaptation to task-specific semantics.

A critical component of prompting, especially for complex reasoning tasks, is the Chain-of-Thought (CoT) \cite{Wei2022ChainOT}. CoT prompting, which encourages the model to generate intermediate reasoning steps, has been shown to significantly improve performance. Research has also focused on automating the generation of these reasoning chains. Auto-CoT \cite{Zhang2022AutomaticCO} clusters questions and generates reasoning chains for each cluster, creating a diverse set of exemplars for few-shot prompting. Shum proposed generating multiple reasoning paths and then selecting the most consistent one to improve robustness\cite{Shum2023AutomaticPA}. Our approach incorporates reward-based learning to optimize prompts that may include CoT-style instructions, but it uniquely integrates a retrieval mechanism to dynamically select the few-shot examples that form the prompt's core.

\subsection{Retrieval-Augmented Language Models}
LLMs possess vast parametric knowledge but it is static and may lack domain-specific or real-time information. Retrieval-Augmented Generation (RAG) was introduced to mitigate this by augmenting prompts with relevant information retrieved from an external knowledge corpus \cite{Lewis2020RetrievalAugmentedGF}. This paradigm has proven effective at reducing factual hallucinations and improving performance on knowledge-intensive tasks.

The synergy between retrieval and in-context learning is particularly relevant to our work. Instead of retrieving factual passages, recent methods retrieve entire exemplars (i.e., input-output pairs) to populate the few-shot context of a prompt. For instance, Ram demonstrated that retrieving examples that are semantically similar to the test input significantly improves few-shot performance\cite{Ram2023InContextRL}. This dynamic selection of examples is more effective than using a fixed, randomly chosen set. These systems typically employ dense retrieval models, such as Sentence-BERT \cite{Reimers2019SentenceBERTSE}, to embed the corpus and query into a shared vector space for similarity search.

Our method, Auto-Prompting with Retrieval Guidance, builds directly on this line of research. However, we extend the paradigm in a crucial way: retrieval is not just a pre-processing step to find examples, but an integral part of the automatic prompt construction loop. We maintain a structured library of prompt components—including instruction templates and domain-specific few-shot examples—and use dense retrieval to dynamically select the most relevant components for each input instance. This allows our system to compose highly specialized prompts tailored to the nuances of hierarchical frame detection in logistics, effectively bridging the gap between automatic prompt optimization and retrieval-augmented in-context learning in a challenging, low-resource industrial setting.

\section{Dataset}
We construct a domain-specific dataset comprising 1,500 text messages collected from a Vietnamese logistics platform. Each instance is a short sentence extracted from real-world customer or internal staff communications. The messages are informal, highly domain-specific, and often include abbreviations, shorthand expressions, and non-standard language structures—reflecting the linguistic challenges typically encountered in operational logistics settings.

Each message is annotated with a three-level hierarchical frame label, capturing the semantic structure of the logistics issue being described:

\begin{itemize}
    \item Level 1 (actor): the primary agent or entity responsible for the issue (e.g., Customer, Shop, Delivery Service, External Factors),
    \item Level 2 (reason): the underlying reason or category of the event (e.g., Incorrect Information, Unavailable, Changed Address, Refused Delivery),
    \item Level 3 (fine-grained cause): a specific instantiation of the reason (e.g., Wrong Product, Out of Stock, Customer on Vacation).
\end{itemize}

The annotation schema includes 73 unique frame labels in total, representing the full cross-product of these three hierarchical levels. The full label space is defined in a structured format such as \foreignlanguage{vietnamese}{Shop – Thay đổi thông tin – Thời gian lấy hàng}, which translates to (Actor=Shop, Reason=Change Info, Detail=Pickup Time).

Annotation was performed manually by domain experts within the logistics company who possess extensive knowledge of business operations and customer communication patterns. All annotations were reviewed and agreed upon through internal consensus to ensure label consistency and semantic fidelity. Notably, the dataset reflects real-world class imbalance: while some frame labels are associated with dozens of instances, others appear only once or twice, creating a naturally low-resource setting for many categories.

This dataset serves as a valuable resource for developing and evaluating frame detection methods under conditions that are both semantically complex and data-scarce, highlighting the practical challenges of applying large language models in real-world industrial domains.

\subsection{Dataset Statistics and Characteristics}

\begin{figure}[!htbp]
    \centering
    \begin{minipage}[t]{0.48\textwidth}
        \centering
        \includegraphics[width=\linewidth]{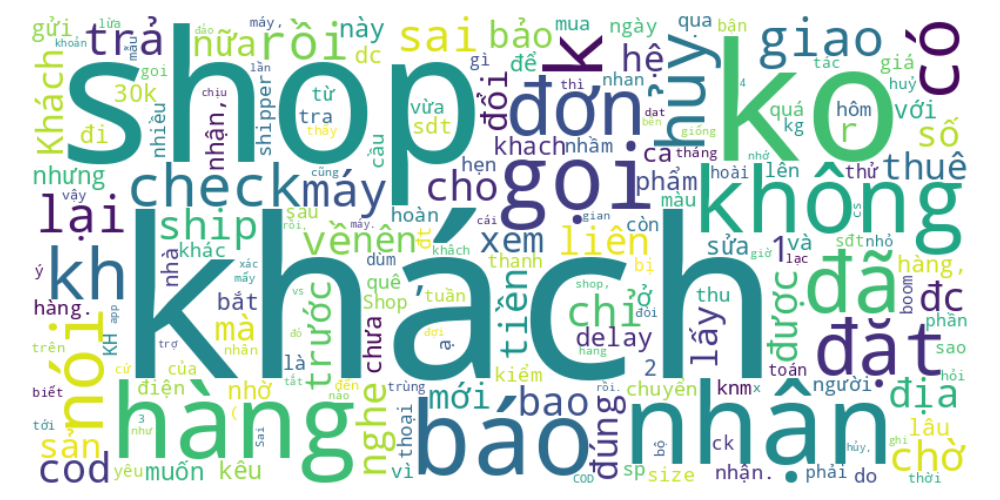}
        \caption{Word cloud of the most frequent tokens in the dataset}
        \label{fig:wordcloud}
    \end{minipage}%
    \hfill
    \begin{minipage}[t]{0.48\textwidth}
        \centering
        \includegraphics[width=\linewidth]{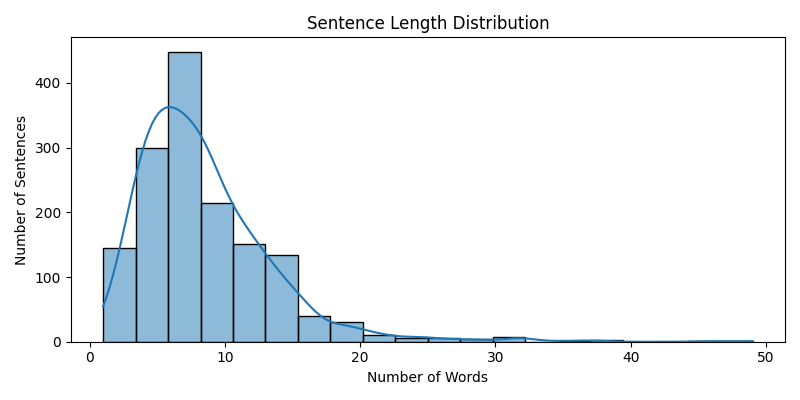}
        \caption{Distribution of sentence lengths (in number of words)}
        \label{fig:sentlen}
    \end{minipage}
\end{figure}

To better understand the linguistic characteristics of the dataset, we conducted descriptive statistical analysis on the first 1,500 annotated messages. The results are summarized in Table~\ref{tab:stats}, while Figure~\ref{fig:wordcloud} and Figure~\ref{fig:sentlen} provide visual insights.

\begin{table}[!hbt]
\centering
\caption{Sentence-level statistics of the annotated dataset}
\label{tab:stats}
\begin{tabular}{|l|r|}
\hline
\textbf{Statistic} & \textbf{Value} \\
\hline
Total number of sentences & 1,500 \\
Average sentence length & 8.48 words \\
Maximum sentence length & 49 words \\
Minimum sentence length & 1 word \\
Sentences longer than 10 words & 393 \\
\hline
Training set (70\%) & 1,050  \\
Validation set (15\%) & 225  \\
Test set (15\%) & 225  \\
\hline
\end{tabular}
\end{table}

From the analysis, we observe that the dataset contains mostly short utterances, with a median around 7–9 words. The vocabulary includes many domain-specific terms such as \textit{shop}, \textit{khách}, \textit{giao}, and informal tokens like \textit{ck}, \textit{kt}, or \textit{sđt}, which reflect both abbreviation and inconsistency in natural language usage. These properties further emphasize the challenges in applying structured learning or fine-tuning methods to the data, and highlight the importance of robust prompting and retrieval mechanisms.
\section{Method}
The loop for generating optimized prompts is repeated three times. Each times we have three prompts using for compare and choose the best prompt.

\begin{figure*}[!hbt]
    \centering
    \includegraphics[width=\linewidth]{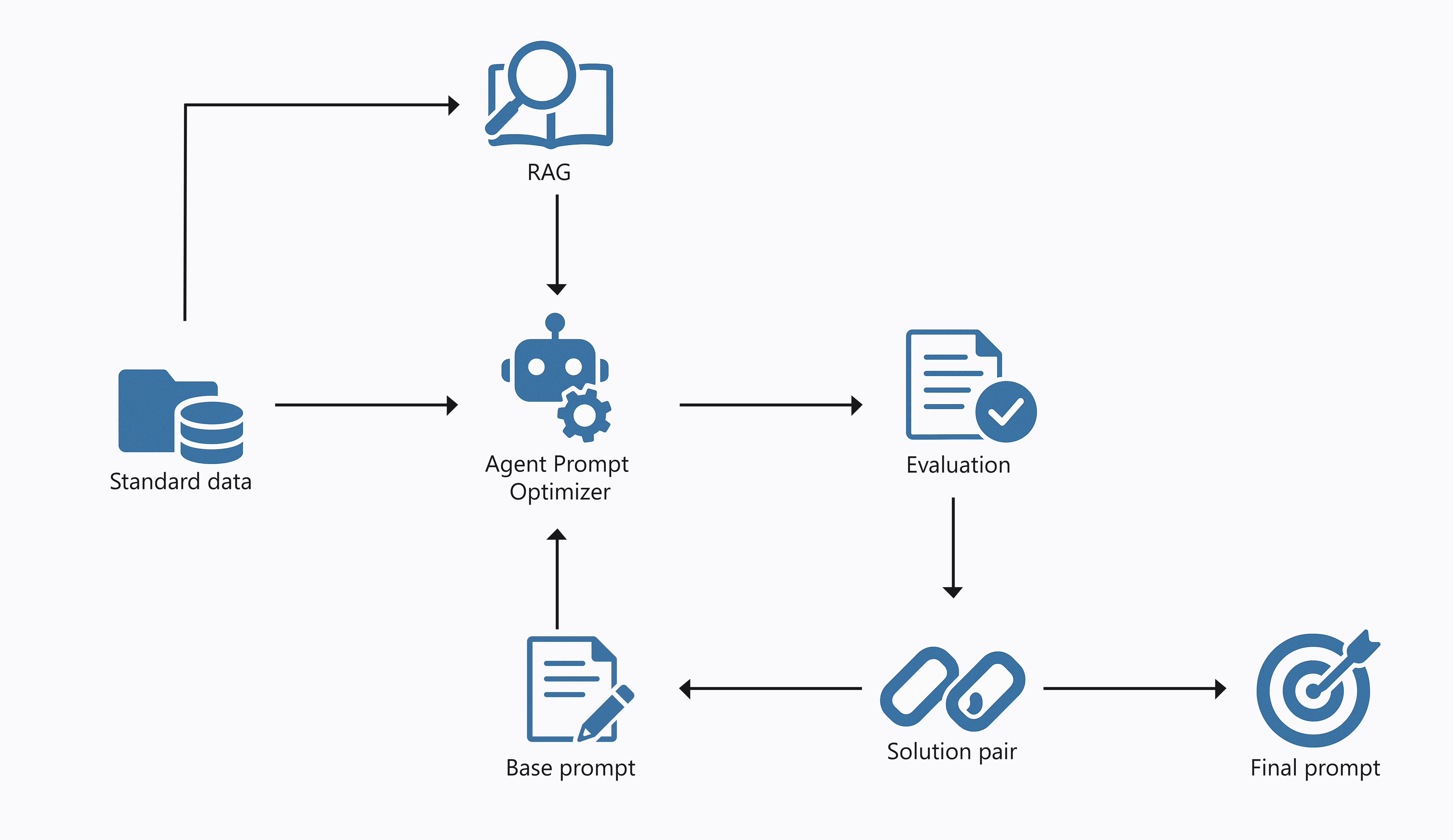}
    \caption{Overview of the proposed prompt optimization pipeline.}
    \label{fig:method_pipeline}
\end{figure*}
To enhance the performance of large language models (LLMs) on text annotation tasks, we propose a prompt optimization pipeline that integrates retrieval-augmented generation (RAG) with an LLM-based prompt optimizer. Our method aims to generate effective task-specific prompts that generalize well during inference. Figure~\ref{fig:method_pipeline} illustrates the architecture of our system.

\subsection{Retrieval-Augmented Generation (RAG)}
To enrich the prompt context with relevant examples, we utilize a dense retrieval mechanism. Each input query is used to retrieve semantically similar examples from a pre-encoded vector database. This process enables the system to dynamically augment the prompt with high-quality, context-relevant exemplars, improving the model's reasoning ability and robustness.

\subsection{Prompt Optimizer Agent}
At the core of our system is the \textit{Prompt Optimizer}, an autonomous agent designed to iteratively improve prompts using a suite of large language models (LLMs). Specifically, we employ a combination of state-of-the-art models, including \textbf{LLaMA 3.1 70B}, \textbf{Qwen 2.5 72B-chat}, \textbf{GPT-4o mini}, and \textbf{GPT-4o}. These models serve both as prompt generators and evaluators, enabling a powerful closed-loop prompt engineering framework.

The optimizer receives as input a \textit{base prompt} and a set of retrieved examples obtained via the RAG module. It applies multiple prompt engineering techniques to generate candidate prompts optimized for the downstream task. Among the key strategies used are:

\begin{itemize}
    \item \textbf{Few-shot prompting}: Retrieved examples are appended to the prompt as demonstrations, allowing the model to learn from specific task instances and better generalize to unseen inputs.
    
    \item \textbf{Chain-of-Thought (CoT) prompting}: Prompts are structured to elicit intermediate reasoning steps before final predictions. This is particularly effective for tasks involving multi-step logic or weak supervision.
    
    \item \textbf{Auto-CoT}: Building upon CoT, we apply Auto-CoT methods to automatically generate rationales from unlabeled data using a two-phase process: (1) reasoning generation with self-consistency sampling, and (2) rationale-filtering based on agreement with ground-truth or simulated outputs. This allows us to bootstrap high-quality CoT-style demonstrations without requiring costly manual annotation.
\end{itemize}

The prompt refinement process operates in a feedback loop. For each generated candidate prompt, we simulate outputs using one or more of the aforementioned LLMs and assess performance using a mix of automated metrics and internal scoring functions. These functions may include comparisons with expert annotations or meta-evaluation via a scoring model (e.g., GPT-4o). Poorly performing prompts are discarded, while promising ones are further refined through transformations such as reordering examples, adjusting instruction specificity, or modifying CoT strategy. Through this iterative optimization process, the Prompt Optimizer acts as a meta-level agent capable of self-evaluating and improving its prompt generation policy. The final prompt selected from this pool is the one that consistently yields the best annotation accuracy and reasoning quality across multiple validation settings. This prompt is then deployed during inference, forming the backbone of the downstream system.

\subsection{Evaluation and Selection}
Each candidate prompt generated by the optimizer is evaluated on a held-out validation set. Evaluation metrics include both automatic measures (e.g., accuracy) and human-in-the-loop assessments performed by domain experts. Prompts are compared in terms of their effectiveness in producing correct annotations.

\subsection{Solution Pairing and Final Selection}
The evaluation phase yields a set of candidate \textit{solution pairs}, where each pair consists of a prompt and its associated performance score. These pairs are ranked and analyzed to determine the best-performing prompt. The selected \textit{Final Prompt} is then used for inference in downstream applications.

The final prompt is deployed for real-world inference scenarios, enabling the LLM to generate accurate and contextually appropriate annotations for new inputs. This prompt encapsulates both task-specific requirements and contextual examples retrieved dynamically during the optimization process.

\section{Experiment}
The main configuration parameters used across all evaluated models are summarized in Table \ref{tab:model-config}.
To ensure consistent reasoning behavior and fair comparison, we fixed the sampling and decoding parameters during all experiments. Specifically, the temperature was set to 0.3 to reduce randomness in generation and encourage deterministic reasoning chains. The top-p (nucleus sampling) and top-k values were set to 0.95 and 70, respectively, balancing diversity and precision in token selection.

\begin{table}[!ht]
\centering
\caption{Model configurations and reasoning capabilities}
\begin{adjustbox}{max width=\columnwidth}
\begin{tabular}{|l|c|c|c|c|}
\hline
\textbf{Parameter} & \textbf{LLaMA 3.1 70B} & \textbf{Qwen 2.5 72B} & \textbf{GPT-4o mini} & \textbf{GPT-4o} \\
\hline
top\_p               & 0.95  & 0.95  & 0.9  & 0.9  \\
top\_k               & 70    & 70    & 50   & 50   \\
temperature          & 0.3   & 0.3   & 0.3  & 0.3  \\
max tokens           & 1024  & 1024  & 1024 & 1024 \\
repetition penalty   & 0     & 0     & 0    & 0    \\
presence penalty     & 0     & 0     & 0    & 0    \\
reasoning            & Yes   & Yes   & Yes  & Yes  \\
\hline
\end{tabular}
\end{adjustbox}
\label{tab:model-config}
\end{table}

We categorize prompting strategies as follows:
\begin{itemize}
    \item \textbf{Manual Prompt (6-shot)}: handcrafted prompt and 6 manually selected examples.
    \item \textbf{Auto Prompt (0-shot)}: prompt automatically generated by the optimizer, no examples.
    \item \textbf{Auto Prompt + RAG (k-shot)}: optimizer-generated prompt with k retrieved in-context examples via semantic retrieval.
\end{itemize}

\begin{table}[!hbt]
\centering
\caption{Comparison of fine-tuning strategies (standard vs. soft prompt) on the test set.}
\begin{adjustbox}{max width=\columnwidth}
\small
\begin{tabular}{|l|l|c|}
\hline
\textbf{Model} & \textbf{Fine-tuning Strategy} & \textbf{Test Accuracy} \\
\hline
DeBERTa-large     & Standard         & 40\% \\
ViDeBERTa-base    & Standard         & 45\% \\
mDeBERTa          & Standard         & 61\% \\
\hline
DeBERTa-large     & Soft Prompt      & 37\% \\
ViDeBERTa-base    & Soft Prompt      & 40\% \\
mDeBERTa          & Soft Prompt      & 57\% \\
\hline
\end{tabular}
\end{adjustbox}
\label{tab:finetuning-strategies}
\end{table}

\begin{table*}[!hbt]
\centering
\caption{Performance of different prompting strategies across models.}
\small
\begin{tabular}{|p{2.5cm}|p{4.5cm}|p{2.cm}|p{2.9cm}|}
\hline
\textbf{Model} & \textbf{Prompting Strategy} & \textbf{Test Accuracy} & \textbf{Real-World Accuracy} \\
\hline
GPT-4o mini      & Manual Prompt (6-shot)         & 83\% & 83\% \\
\hline
GPT-4o mini      & Auto Prompt (0-shot)           & 82\% & 80\% \\
\hline
GPT-4o mini      & Auto Prompt + RAG (3-shot)     & 85\% & 84\% \\
\hline
\textbf{GPT-4o mini} & \textbf{Auto Prompt + RAG (6-shot)} & \textbf{88\%} & \textbf{88\%} \\
\hline
GPT-4o           & Manual Prompt (6-shot)         & 84\% & 84\% \\
\hline
GPT-4o           & Auto Prompt (0-shot)           & 84\% & 84\% \\
\hline
GPT-4o           & Auto Prompt + RAG (3-shot)     & 86\% & 88\% \\
\hline
\textbf{GPT-4o}  & \textbf{Auto Prompt + RAG (6-shot)} & \textbf{90\%} & \textbf{92\%} \\
\hline
LLaMA 3.1 70B    & Manual Prompt (6-shot)         & 72\% & 70\% \\
\hline
LLaMA 3.1 70B    & Auto Prompt (0-shot)           & 74\% & 80\% \\
\hline
LLaMA 3.1 70B    & Auto Prompt + RAG (3-shot)     & 79\% & 81\% \\
\hline
\textbf{LLaMA 3.1 70B} & \textbf{Auto Prompt + RAG (6-shot)} & \textbf{87\%} & \textbf{87\%} \\
\hline
Qwen 2.5 72B     & Manual Prompt (6-shot)         & 73\% & 71\% \\
\hline
Qwen 2.5 72B     & Auto Prompt (0-shot)           & 76\% & 80\% \\
\hline
Qwen 2.5 72B     & Auto Prompt + RAG (3-shot)     & 79\% & 81\% \\
\hline
\textbf{Qwen 2.5 72B} & \textbf{Auto Prompt + RAG (6-shot)} & \textbf{87\%} & \textbf{87\%} \\
\hline
\end{tabular}
\label{tab:model-comparison}
\end{table*}

Table~\ref{tab:model-comparison} summarizes performance across several LLMs under varying prompting strategies. We evaluate zero-shot, manual few-shot, and our proposed auto-prompting with retrieval-augmented few-shot examples. The goal is to assess how models adapt with optimized prompts selected dynamically based on input similarity.

\paragraph{Auto Prompt + RAG is Most Effective.}
Across all models, the best results are consistently obtained using our proposed 6-shot Auto Prompt + RAG strategy. For example, \textbf{GPT-4o} reaches \textbf{90\%} test and \textbf{92\%} real-world accuracy, outperforming both manual and zero-shot prompts. This supports the effectiveness of combining LLM-based prompt optimization with retrieval-guided example selection.

\paragraph{Manual vs. Retrieval-based Prompting.}
Manual 6-shot prompts generally underperform. For instance, GPT-4o mini improves from 83\% (manual) to 88\% (RAG-6-shot), and similar gains are seen with Qwen and LLaMA. This shows that automatic retrieval often selects more relevant demonstrations than handcrafted examples.

\paragraph{Model Trends.}
GPT-4o and its mini variant show strong, consistent performance. In contrast, Qwen and LLaMA start lower in zero-shot (74–76\%) but improve markedly with retrieval-augmented prompts (up to 87\%). This highlights the value of prompt quality over raw model size.

\paragraph{Real-World Generalization.}
High test accuracy correlates well with real-world performance. Models like GPT-4o and Qwen retain or improve accuracy post-deployment. Slight drops in LLaMA suggest generalization may depend on alignment and robustness to input variation.

\paragraph{Fine-tuning Baselines.}
Table~\ref{tab:finetuning-strategies} compares standard and soft prompt fine-tuning on encoder-based models. Even the best result (57\% for mDeBERTa with soft prompt) lags far behind prompting-based LLMs. This reinforces the strength of in-context learning with optimized prompting pipelines over traditional fine-tuning approaches.

\paragraph{Conclusion.}
Our findings show that combining automatic prompt generation with retrieval-augmented few-shot examples offers a scalable and effective alternative to fine-tuning. It improves both performance and generalization, while being model-agnostic and parameter-efficient.

\section{Limitations}

While our method achieves strong performance in both controlled and real-world settings, it has several notable limitations:

\begin{itemize}
    \item \textbf{Domain specificity:} Evaluation is confined to the \textbf{logistics domain}, which, while enabling precise alignment, limits \textbf{generalizability} to other domains or languages.

    \item \textbf{Evaluation subjectivity:} Real-world performance is assessed by \textbf{domain experts}, not NLP researchers, introducing \textbf{subjectivity} and lacking formal metrics for reasoning depth or faithfulness.

    \item \textbf{RAG sensitivity to informal input:} The retrieval component struggles with \textbf{slang, shorthand, or non-standard phrasing}, leading to \textbf{suboptimal few-shot selection}.

    \item \textbf{Fixed iteration prompting:} The prompt optimizer performs a \textbf{fixed number of refinement steps}, which may \textbf{fail to converge} and adds \textbf{latency}, limiting \textbf{real-time or large-scale applicability}.

    \item \textbf{Compute requirements:} High-performing models (e.g., \textbf{GPT-4o, Qwen 72B}) require substantial \textbf{compute resources}, which is impractical for \textbf{low-resource or edge deployment} scenarios.

    \item \textbf{Limited label diversity:} Training data often includes only \textbf{1–2 examples per class}, which reduces the system’s ability to select \textbf{representative demonstrations}, limiting \textbf{few-shot generalization} even with RAG support.
\end{itemize}

\bibliographystyle{ieeetran1}
\bibliography{custom}

\appendix

\section{Example Prompt} \label{appendix:prompts}

In this appendix, we present the prompt templates that were employed in the logistics frame detection task. These prompts were designed to guide large language models (LLMs) such as GPT-4o, Qwen 2.5, and LLaMA 3.1 in generating preliminary annotations. We experimented with zero-shot, few-shot, chain-of-thought, automatic chain-of-thought (Auto-CoT), and retrieval-augmented prompts.  

\subsection{Zero-shot Prompt}
\begin{framed}
\textbf{Zero-shot Frame Detection Prompt}  

\textbf{User:}  
Identify the frame structure in the following logistics message.  
The frame label has three levels:  
\begin{itemize}
    \item Actor (Customer, Shop, Delivery Service, External Factor)  
    \item Reason (Incorrect Information, Unavailable, Changed Address, Refused Delivery, …)  
    \item Fine-grained Cause (e.g., Wrong Product, Out of Stock, Customer on Vacation)  
\end{itemize}

Text: ``<input logistics message>''  

\textbf{Assistant:}  
\{ "actor": ..., "reason": ..., "cause": ... \}  
\end{framed}

\subsection{Few-shot Prompt}
\begin{framed}
\textbf{Few-shot Frame Detection Prompt}  

\textbf{User:}  
Here are examples of annotated logistics messages. Please follow the same format for the new message.  

Example 1:  
Text: ``\begin{otherlanguage}{vietnamese}Khách không nhận vì giao trễ.\end{otherlanguage}''  
Output: ("Customer", "Refused Delivery", "Late Delivery")  

Example 2:  
Text: ``\begin{otherlanguage}{vietnamese}Shop báo sai địa chỉ nhận.\end{otherlanguage}''  
Output: ("Shop", "Incorrect Information", "Wrong Address")  

Now annotate the following text:  
``<input logistics message>''  

\textbf{Assistant:}  
("...", "...", "...")  
\end{framed}

\subsection{Chain-of-Thought Prompt}
\begin{framed}
\textbf{Chain-of-Thought Frame Detection Prompt}  

\textbf{User:}  
Analyze the logistics message step by step.  
\begin{enumerate}
    \item Identify the main actor.  
    \item Identify the reason category.  
    \item Determine the fine-grained cause.  
\end{enumerate}
Finally, return the structured frame label.  

Text: ``<input logistics message>''  

\textbf{Assistant:}  
Step 1: Actor = ...  
Step 2: Reason = ...  
Step 3: Cause = ...  
Final Output: \{ "actor": ..., "reason": ..., "cause": ... \}  
\end{framed}

\subsection{Auto-CoT Prompt}
\begin{framed}
\textbf{Automatic Chain-of-Thought Prompt}  

\textbf{User:}  
``<input logistics message>''  

\textbf{Assistant:}  
Reasoning Path A: ... $\rightarrow$ Candidate frame label.  
Reasoning Path B: ... $\rightarrow$ Candidate frame label.  
Reasoning Path C: ... $\rightarrow$ Candidate frame label.  

\textbf{User:}  
Select the most consistent reasoning path and output the final annotation.  

\textbf{Assistant:}  
\{ "actor": ..., "reason": ..., "cause": ... \}  
\end{framed}

\subsection{Retrieval-Augmented Prompt}
\begin{framed}
\textbf{RAG-Enhanced Few-shot Prompt}  

\textbf{User:}  
Given the following retrieved examples from similar logistics cases, use them to annotate the new message.  

Retrieved Example 1:  
Text: ``\begin{otherlanguage}{vietnamese}
Khách đi vắng nên không nhận hàng.
\end{otherlanguage}''  
Output: ("Customer", "Unavailable", "On Vacation")  

Retrieved Example 2:  
Text: ``\begin{otherlanguage}{vietnamese}Giao hàng thất bại do thiếu số điện thoại.\end{otherlanguage}''  
Output: ("Delivery Service", "Incorrect Information", "Missing Contact Info")  

New text: ``<input logistics message>''  

\textbf{Assistant:}  
("...", "...", "...")  
\end{framed}
\subsection{Evaluation-Refinement Prompt}
\begin{framed}
\textbf{Evaluation and Refinement Prompt}  

\textbf{User:}  
You previously used the following prompt for frame detection:  
<previous prompt text>  

Here are some annotated cases where the model prediction was incorrect:  \\
Case 1:  ...\\
Case 2:  ...

\textbf{Assistant:}  
Step 1: Analyze the error in each case and explain why the prediction was wrong.  
Step 2: Identify missing or ambiguous instructions in the current prompt.  
Step 3: Propose modifications to improve the prompt.  
Step 4: Output a refined version of the prompt.  

\textbf{Assistant (Example Reasoning):}  
- Case 1: The model confused ``Out of Stock'' with ``Wrong Product'' because the prompt did not emphasize distinguishing product mismatch from stock issues.  
- Case 2: The model misclassified the actor. The prompt should explicitly instruct to check who is responsible (Customer vs. Delivery Service).  

Refined Prompt:  
``When detecting frames, always check \\(1) who is responsible for the issue (actor), \\(2) whether the problem comes from product mismatch vs. stock shortage \\(3) if the delivery service failed due to missing information. Return output as {actor, reason, cause}.''  
\end{framed}

\subsection{Multi-Agent Debate and Self-Improvement Prompt}
\begin{framed}
\textbf{Multi-Agent Debate Prompt}  

\textbf{User:}  
You are Agent A. Given the following logistics message, provide your frame detection output:  
Text: ...
\textbf{Assistant (Agent A):}  
("Customer", "Refused Delivery", "Late Delivery")  
\textbf{User:}  
Now here are the outputs from other agents:  
- Agent B: ("Delivery Service", "Unavailable", "Late Delivery")  
- Agent C: ("Customer", "Changed Address", "Wrong Time")  
Using these alternative answers, explain step by step which reasoning is more consistent, identify weaknesses in your own answer, and propose an improved solution.  

\textbf{Assistant (Agent A Revised):}  
Step 1: Compare outputs: Agent B incorrectly assigns the actor; Agent C mislabels the reason.  
Step 2: My original output is mostly correct, but I need to emphasize actor detection rules.  
Step 3: Improved answer: ("Customer", "Refused Delivery", "Late Delivery")  

\end{framed}

\begin{framed}
\textbf{Self-Improvement Debate Prompt}  

\textbf{User:}  
Multiple agents have generated candidate prompts for frame detection. Here are three prompt variants:  

Prompt A: ``Identify actor, reason, and cause. Return as JSON.''  
Prompt B: ``Step by step, decide actor first, then reason, then cause.''  
Prompt C: ``Focus on who is responsible, why it happened, and the fine-grained detail.''  

Debate with each other to decide which prompt is most effective. Consider coverage, clarity, and ability to prevent errors seen in past cases. After the debate, propose a new improved prompt that combines the strengths of A, B, and C.  

\textbf{Assistant (Consensus):}  
- Agent 1: Prompt A is too general, may cause confusion.  
- Agent 2: Prompt B enforces reasoning steps, but might be verbose.  
- Agent 3: Prompt C adds clarity on actor responsibility.  

Final Improved Prompt:  
``When analyzing a logistics message:  
1. Identify who is responsible (actor).  
2. Identify the general reason category.  
3. Identify the fine-grained cause.  
Always distinguish product mismatch vs. stock issues, and customer refusal vs. delivery service error. Return output in JSON format as {"actor": ..., "reason": ..., "cause": ...}.''  
\end{framed}

\end{document}